\documentclass{article}

\usepackage{arxiv}
\usepackage[numbers]{natbib}

\usepackage[utf8]{inputenc} 
\usepackage[T1]{fontenc}    
\usepackage{hyperref}       
\usepackage{url}            
\usepackage{booktabs}       
\usepackage{amsfonts}       
\usepackage{nicefrac}       
\usepackage{microtype}      
\usepackage{lipsum}		
\usepackage{graphicx}
\usepackage{natbib}
\usepackage{doi}
\usepackage{subcaption}

\title{Self-Supervised Image Representation Learning: Transcending Masking with Paired Image Overlay}

\author{
    \begin{tabular}{c c c}
    Yinheng Li & Han Ding & Shaofei Wang\\
    Columbia University & Columbia University & Columbia University\\
    \texttt{yl4039@columbia.edu} & \texttt{hd2412@columbia.edu} & \texttt{sw3316@columbia.edu}
    \end{tabular}
}

\hypersetup{
pdftitle={Overlaying},
pdfauthor={Yinheng Li},
pdfkeywords={Self-supervised Learning, Autoencoder, Augmentation, pairwise training}
}

\begin{document}
\maketitle

\begin{abstract}
Self-supervised learning has become a popular approach in recent years for its ability to learn meaningful representations without the need for data annotation. This paper proposes a novel image augmentation technique, overlaying images, which has not been widely applied in self-supervised learning. This method is designed to provide better guidance for the model to understand underlying information, resulting in more useful representations. The proposed method is evaluated using contrastive learning, a widely used self-supervised learning method that has shown solid performance in downstream tasks. The results demonstrate the effectiveness of the proposed augmentation technique in improving the performance of self-supervised models.

\end{abstract}


\section{Introduction}

Self-supervised learning has seen significant success in recent years in both natural language processing and computer vision domains, particularly with the advent of attention mechanisms and large language models such as BERT and GPT. While transformer-based models are the focus in the NLP community for self-supervised learning, computer vision research has employed a variety of approaches. One trend is the use of contrastive learning with data augmentation, while another trend is to adapt the idea of masked training from NLP to computer vision.

However, applying masked training to images poses challenges. Unlike text, images have high dimensions and spatial relationships, making it difficult to mask a single pixel and resulting in large attention matrices. As a result, researchers have divided images into patches and treated each patch as the smallest unit, as seen in works such as \cite{dosovitskiy2020image, he2021masked}.

In this paper, we propose a new method for image corruption, which is closer to the idea of "masking" and aims to better leverage masked training in computer vision tasks. Our method is to overlay one image onto another, a technique that has been implemented by \cite{inoue2018data} and shown to be effective. To the best of our knowledge, this technique has not been used in self-supervised learning. We demonstrate the effectiveness of this method in comparison to existing augmentation techniques through experiments on self-supervised models.

\begin{figure}[h!]
  \includegraphics[scale=0.6]{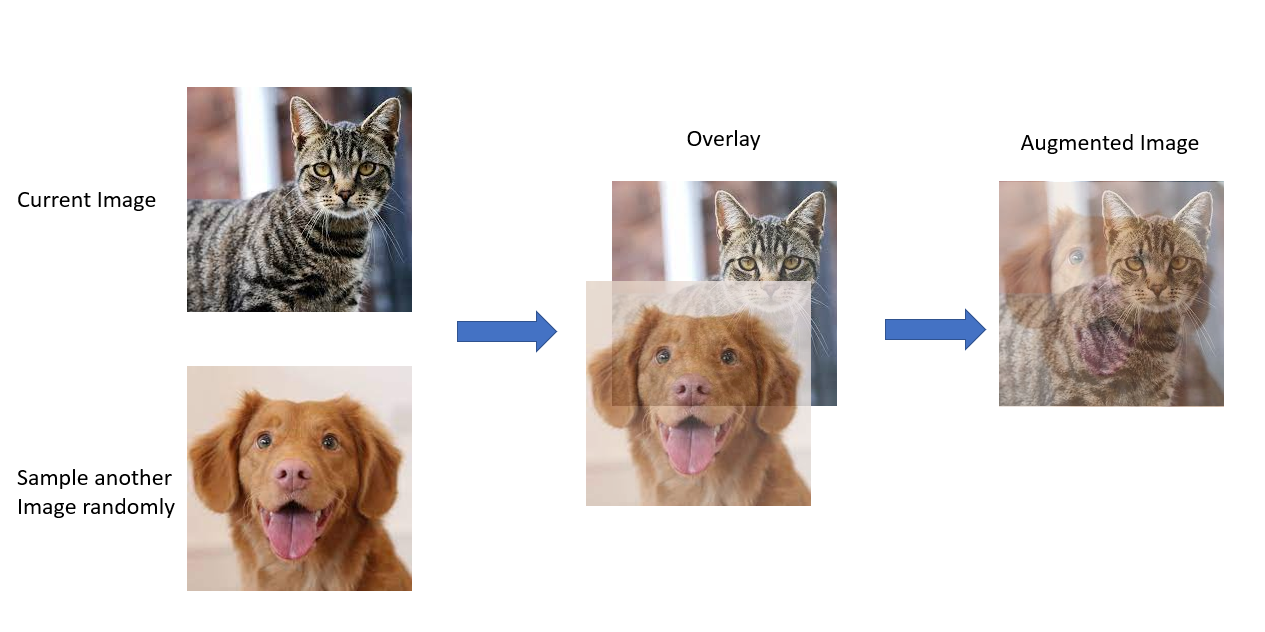}
  \caption{proposed transformation}
\end{figure}

\section{Related Work}

\textbf{Autoencoders} are a traditional self-supervised learning algorithm that trains to reconstruct the input from a learned low-dimension representation. Variants of autoencoders include denoising autoencoders \cite{vincent2010stacked}, where noise is added to the input during training. Our proposed method can be viewed as a special type of denoising autoencoder, where we use image overlap to create a corrupted image. Recently, self-supervised learning using autoencoders for computer vision tasks has also achieved great success \cite{he2021masked}.

\textbf{Data augmentation} is a widely used technique in computer vision research, with numerous techniques such as flipping, cropping, and coloring (\cite{shorten2019survey}). Image overlay (also known as image mixture) is also a type of augmentation method, but has been less explored compared to others.

\section{Approach}

The proposed method is based on the idea of image overlay, which is a form of image corruption, similar to image masking. There are several reasons why image overlay is a suitable method for self-supervised representation learning in computer vision.

First, image overlay is conceptually similar to image masking. When a patch of an image is masked, it can be viewed as overlaying an image mask onto the image, which is a special case of overlaying one image on top of another.

Second, image overlay provides better guidance for self-supervised representation learning compared to standard masking. In masked image training, the model is asked to reconstruct the missing part of the image, which is a difficult task in the domain of vision due to the high dimension of an image and the uncertainty of the real world. For example, predicting what is behind a window is a challenging task even for humans, due to the infinite possibilities. On the other hand, distinguishing between two films overlapped with one another is a relatively easy task as long as one has certain knowledge of the world. Therefore, training a model to distinguish and recover the overlapped image demands the model to pick up knowledge about the object and the world without worrying about uncertainty.

Third, the method of image overlay is also similar to the idea of constructive learning, where the model needs to distinguish between positive and negative pairs. In this method, the model is also trained to distinguish between two samples, except they are in the same image.

Lastly, this method is straightforward to implement as it does not require constructing any complex transformations on the image or sampling positive and negative pairs. The only requirement is to change the transparency of an image and overlay it onto another image.

\begin{figure}[h!]
  \includegraphics[scale=0.65]{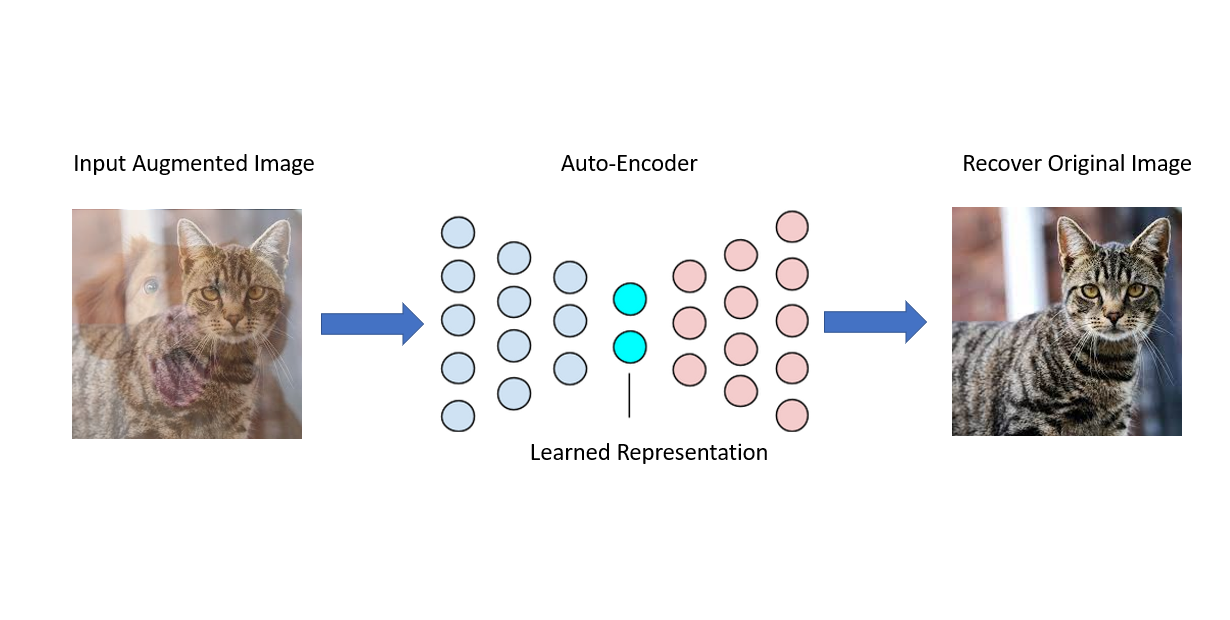}
  \caption{training pipeline}
\end{figure}
\subsection{Pretraining}
A detailed approach is shown in Figure2. Denote input image $x_i$, current batch $X_b$, encoder $f_e(x)$, decoder $f_d(x)$

\textbf{Step1}: random sample image $x_j$ from $X_b$

\textbf{Step2}: generate an augmented image $\hat{x_i} = \alpha x_j + (1-\alpha) x_i$. $\alpha$ is a hyperparameter which controls the transparency of the image overlap. It should be less than 0.5 as the input image $x_i$ should dominates the information in the picture.  

\textbf{Step3}: train the model to minimize the loss $L(x_i) = MSE(f_d(f_e(\hat{x_i})), x_i)$. MSE is short for mean squared error.

Once the model is trained, we use $f_e$ as the encoder to extract meaningful representation from the image. The performance will be evaluated in downstream tasks such as image classification.

\subsection{Autoencoder}
We use a Resnet50 architecture (\cite{he2016deep}) as the backbone for the autoencoder, following the implementation in \cite{Horizon2333github}. The encoder is composed of Resnet50 and the decoder is a reversed version of Resnet50.

\subsection{Downstream Evaluation}
To evaluate the performance of the pretrained model, the encoder is fine-tuned on a downstream task. In our experiment, we use image classification as the downstream task. A linear head is added to the encoder for classification purposes. During downstream evaluation and testing, no additional augmentation is applied to the input.

\section{Object256 Experiment}
We conduct our experiments on the Object256 dataset (\cite{griffin2007caltech}), which consists of 29,780 images covering 256 objects. The dataset is split into 23,824 images for the training set and 5,956 images for the testing set. In the pretraining phase, we train the autoencoder with image overlay $f^{\alpha}$, where $\alpha \in{0, 0.1, 0.2, 0.3, 0.4}$ and $f^{\alpha} = f_d(f_e(\hat{X}))$. Additional configuration details can be found in Table a. Our data loader's input transformation includes (1) image overlay with $\alpha$ ($\alpha=0$ means no overlay) (2) random resized crop (to ensure the input image is of size $244\times 244$). To compare our augmentation method with other commonly used methods, we also trained two sets of models with random mask and AutoAugment (auto-policy) transformation: $f^{random mask}$ and $f^{auto policy}$. The random mask transformation randomly sets $m\%$ (where $m\in{\{1,2,3,4}\}$) of pixels in an image to 0 (blank), which is equivalent to adding white noise to the input, as shown in Figure 3. AutoAugment (\cite{cubuk2019autoaugment}) is a set of augmentations that were optimized on the Imagenet dataset.

\begin{figure}[h!]
  \includegraphics[scale=0.6]{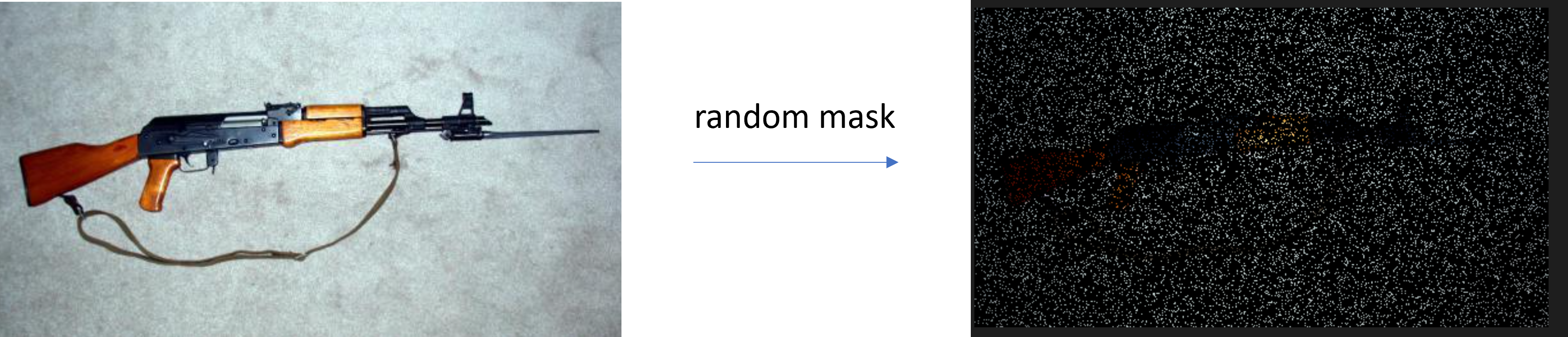}
  \caption{an example of random mask transformation}
\end{figure}

\begin{figure}[h!]
  \includegraphics[scale=0.4]{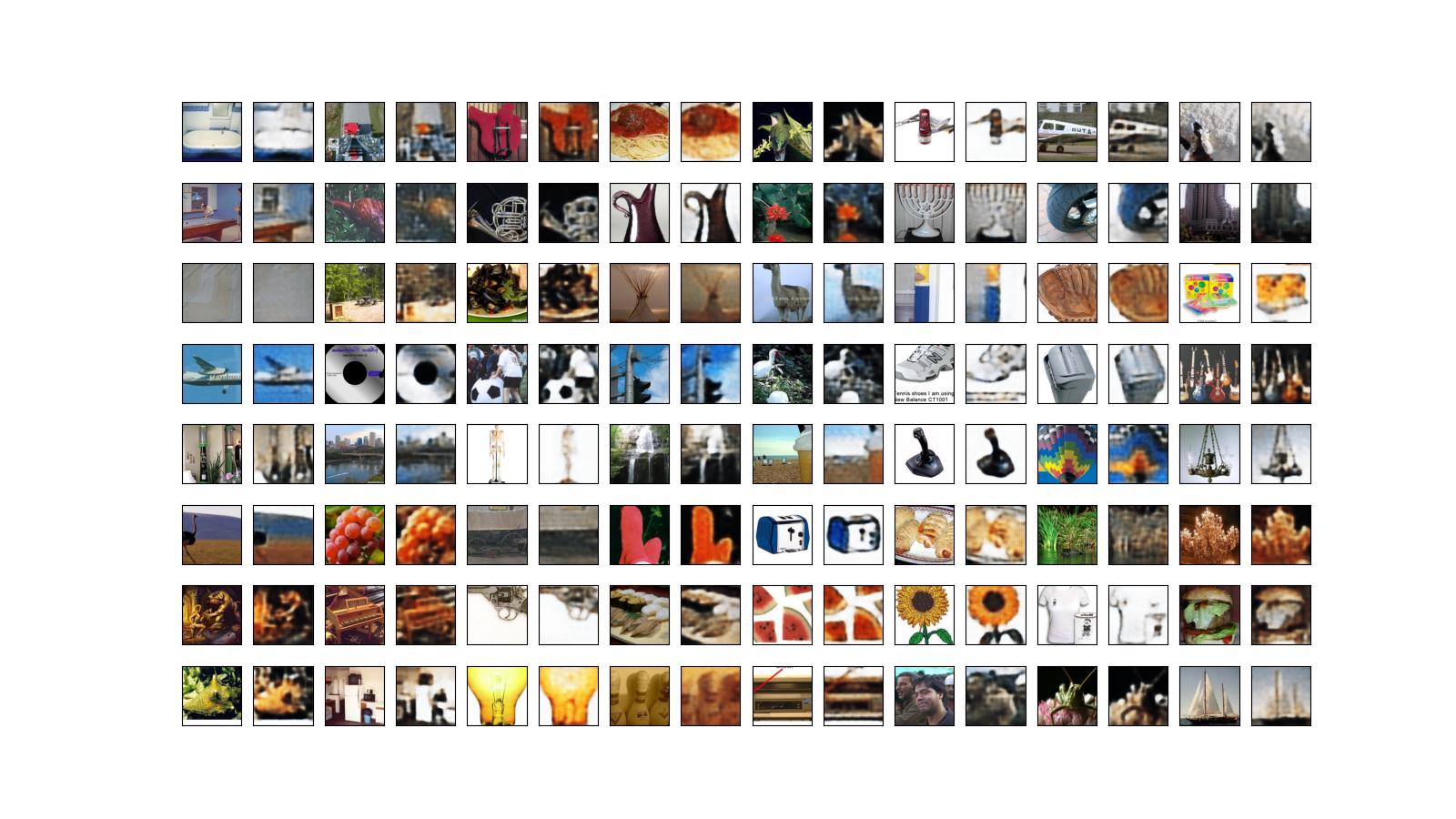}
  \caption{This is the reconstruction result using $f^{\alpha=0.2}$. We only show one result because results from different $\alpha$'s are visually similar to this picture.}
\end{figure}

Figure 4 shows the reconstruction results on 64 random samples from the testing set. All $\alpha$ configurations show very similar reconstruction results, indicating that $f^{\alpha}$ is sufficiently trained.

We then use the pretrained encoder $f^{\alpha}_e$ and fine-tune it on the downstream image classification task using the obj256 training set for fine-tuning and the test set for evaluation. In the fine-tuning phase, each encoder is trained for 20 epochs without any additional transformation. Configuration details can be found in Table b. The best accuracy on the testing set over 20 epochs is reported. We also train a Resnet model without pretraining for 20 epochs as a baseline for comparison.

\begin{figure}[h]
    \centering
    \begin{subfigure}[b]{0.45\textwidth}
        \begin{tabular}{ll}
        config        & value  \\
        \hline
        epoch         & 50     \\
        learning rate & 0.05   \\
        momentum      & 0.9    \\
        optimizer     & SGD    \\
        batch size    & 256    \\
        scheduler     & cosine \\
        weight decay  & 1e-4  \\
        \end{tabular}
        \caption{model configuration-pretrain}
    \end{subfigure}
    \begin{subfigure}[b]{0.45\textwidth}
        \begin{tabular}{ll}
        config        & value  \\
        \hline
        epoch         & 20     \\
        learning rate & 0.01   \\
        momentum      & 0.9    \\
        optimizer     & SGD    \\
        batch size    & 256    \\
        scheduler     & cosine \\
        weight decay  & 1e-4  \\
        \end{tabular}
        \caption{model configuration-finetune}
    \end{subfigure}
\end{figure}

The first experiment is to compare the model's performance on downstream classification task with different $\alpha$s. According to table 1, the accuracy across different $\alpha$ is around 37\%. This accuracy looks very low given a resnet50 model can easily achieve 80\% on imagenet with one thousand class. However, we looked up other people's experiment and confirmed that this is a reasonable result given our architecture. More results can be found from this website [\cite{xufanxionggithub}]. According to their experiment, the top performer without transferred learning is only 39.06\% . The reason that all models have a low accuracy on this dataset is that this is a small dataset with a large number of classes. 

The best model in table3 under finetuning is a pre-trained encoder with $\alpha = 0.3$. We also did linear probing where we freeze all layers except the last layer for classification and trained for 5 epochs. But we found all models have very similar performance during linear probing. 

Lastly, we compared the performance of using image overlap, using random masking, supervised learning without pretraining and auto-policy augmented pretraining. Results are shown in table 4. We found the model with image overlap augmentation still performs the best among all the other models. 

\begin{table}[]
\centering
\begin{tabular}{lll}
model        & fine-tune-accuracy  & linear-probing-accuracy  \\
\hline
$\alpha=0$   & 37.1\% & 7.7\%   \\
$\alpha=0.1$ & 37.0\% & 7.6\%   \\
$\alpha=0.2$ & 35.5\% & 7.7\%   \\
$\alpha=0.3$ & \textbf{38.1\%}  & 7.6\%  \\
$\alpha=0.4$ & 37.0\%  & 7.6\%  \\
\end{tabular}
\caption{accuracy with different alpha}
\end{table}

\begin{table}[h!]
\centering
\begin{tabular}{ll}
model                & image classification accuracy  \\
\hline
best-image-overlap   & \textbf{38.1\%}    \\
best-image-masking      & 36.7\%    \\
supervised-learning  & 32.5\%    \\
auto-policy augmentation & 33.2\%    \\
\end{tabular}
\caption{accuracy with different models}
\end{table}

\section{Discussion and Conclusion}
In this paper, we proposed a new method of image "masking" using image overlay for self-supervised learning in computer vision. We showed that this method is simple and easy to implement, and it outperforms other commonly used self-supervised methods on the Object256 dataset. However, it is important to note that the Object256 dataset is relatively small and the classification accuracy is generally low across all models. Therefore, more extensive experimentation on larger datasets such as Imagenet is needed to fully evaluate the effectiveness of our proposed method.

Additionally, we only used a Resnet architecture in our experiments due to computational constraints. Therefore, another direction for future work is to test our method using transformer-based neural networks.

Overall, our proposed method of image overlay for self-supervised learning in computer vision provides a promising approach to learn meaningful representations without the need for data annotation. It is simple to implement, and we hope that future research will further explore its potential.

\bibliographystyle{ieeetr}

\bibliography{ref} 

\begin{thebibliography}{17}
\providecommand{\natexlab}[1]{#1}
\providecommand{\url}[1]{\texttt{#1}}
\expandafter\ifx\csname urlstyle\endcsname\relax
  \providecommand{\doi}[1]{doi: #1}\else
  \providecommand{\doi}{doi: \begingroup \urlstyle{rm}\Url}\fi

\bibitem[Vaswani et~al.(2017)Vaswani, Shazeer, Parmar, Uszkoreit, Jones, Gomez,
  Kaiser, and Polosukhin]{vaswani2017attention}
Ashish Vaswani, Noam Shazeer, Niki Parmar, Jakob Uszkoreit, Llion Jones,
  Aidan~N Gomez, {\L}ukasz Kaiser, and Illia Polosukhin.
\newblock Attention is all you need.
\newblock \emph{Advances in neural information processing systems}, 30, 2017.

\bibitem[Devlin et~al.(2018)Devlin, Chang, Lee, and Toutanova]{devlin2018bert}
Jacob Devlin, Ming-Wei Chang, Kenton Lee, and Kristina Toutanova.
\newblock Bert: Pre-training of deep bidirectional transformers for language
  understanding.
\newblock \emph{arXiv preprint arXiv:1810.04805}, 2018.

\bibitem[Brown et~al.(2020)Brown, Mann, Ryder, Subbiah, Kaplan, Dhariwal,
  Neelakantan, Shyam, Sastry, Askell, et~al.]{brown2020language}
Tom Brown, Benjamin Mann, Nick Ryder, Melanie Subbiah, Jared~D Kaplan, Prafulla
  Dhariwal, Arvind Neelakantan, Pranav Shyam, Girish Sastry, Amanda Askell,
  et~al.
\newblock Language models are few-shot learners.
\newblock \emph{Advances in neural information processing systems},
  33:\penalty0 1877--1901, 2020.

\bibitem[Chen et~al.(2020)Chen, Kornblith, Norouzi, and Hinton]{chen2020simple}
Ting Chen, Simon Kornblith, Mohammad Norouzi, and Geoffrey Hinton.
\newblock A simple framework for contrastive learning of visual
  representations.
\newblock In \emph{International conference on machine learning}, pages
  1597--1607. PMLR, 2020.

\bibitem[He et~al.(2020)He, Fan, Wu, Xie, and Girshick]{he2020momentum}
Kaiming He, Haoqi Fan, Yuxin Wu, Saining Xie, and Ross Girshick.
\newblock Momentum contrast for unsupervised visual representation learning.
\newblock In \emph{Proceedings of the IEEE/CVF conference on computer vision
  and pattern recognition}, pages 9729--9738, 2020.

\bibitem[Caron et~al.(2020)Caron, Misra, Mairal, Goyal, Bojanowski, and
  Joulin]{caron2020unsupervised}
Mathilde Caron, Ishan Misra, Julien Mairal, Priya Goyal, Piotr Bojanowski, and
  Armand Joulin.
\newblock Unsupervised learning of visual features by contrasting cluster
  assignments.
\newblock \emph{Advances in Neural Information Processing Systems},
  33:\penalty0 9912--9924, 2020.

\bibitem[Dosovitskiy et~al.(2020)Dosovitskiy, Beyer, Kolesnikov, Weissenborn,
  Zhai, Unterthiner, Dehghani, Minderer, Heigold, Gelly,
  et~al.]{dosovitskiy2020image}
Alexey Dosovitskiy, Lucas Beyer, Alexander Kolesnikov, Dirk Weissenborn,
  Xiaohua Zhai, Thomas Unterthiner, Mostafa Dehghani, Matthias Minderer, Georg
  Heigold, Sylvain Gelly, et~al.
\newblock An image is worth 16x16 words: Transformers for image recognition at
  scale.
\newblock \emph{arXiv preprint arXiv:2010.11929}, 2020.

\bibitem[He et~al.(2021)He, Chen, Xie, Li, Doll{\'a}r, and
  Girshick]{he2021masked}
Kaiming He, Xinlei Chen, Saining Xie, Yanghao Li, Piotr Doll{\'a}r, and Ross
  Girshick.
\newblock Masked autoencoders are scalable vision learners.
\newblock \emph{arXiv preprint arXiv:2111.06377}, 2021.

\bibitem[Misra and Maaten(2020)]{misra2020self}
Ishan Misra and Laurens van~der Maaten.
\newblock Self-supervised learning of pretext-invariant representations.
\newblock In \emph{Proceedings of the IEEE/CVF Conference on Computer Vision
  and Pattern Recognition}, pages 6707--6717, 2020.

\bibitem[Inoue(2018)]{inoue2018data}
Hiroshi Inoue.
\newblock Data augmentation by pairing samples for images classification.
\newblock \emph{arXiv preprint arXiv:1801.02929}, 2018.

\bibitem[Vincent et~al.(2010)Vincent, Larochelle, Lajoie, Bengio, Manzagol, and
  Bottou]{vincent2010stacked}
Pascal Vincent, Hugo Larochelle, Isabelle Lajoie, Yoshua Bengio, Pierre-Antoine
  Manzagol, and L{\'e}on Bottou.
\newblock Stacked denoising autoencoders: Learning useful representations in a
  deep network with a local denoising criterion.
\newblock \emph{Journal of machine learning research}, 11\penalty0 (12), 2010.

\bibitem[Shorten and Khoshgoftaar(2019)]{shorten2019survey}
Connor Shorten and Taghi~M Khoshgoftaar.
\newblock A survey on image data augmentation for deep learning.
\newblock \emph{Journal of big data}, 6\penalty0 (1):\penalty0 1--48, 2019.

\bibitem[He et~al.(2016)He, Zhang, Ren, and Sun]{he2016deep}
Kaiming He, Xiangyu Zhang, Shaoqing Ren, and Jian Sun.
\newblock Deep residual learning for image recognition.
\newblock In \emph{Proceedings of the IEEE conference on computer vision and
  pattern recognition}, pages 770--778, 2016.

\bibitem[Horizon2333(2022)]{Horizon2333github}
Horizon2333.
\newblock imagenet-autoencode, 2022.
\newblock URL \url{https://github.com/Horizon2333/imagenet-autoencoder}.

\bibitem[Griffin et~al.(2007)Griffin, Holub, and Perona]{griffin2007caltech}
Gregory Griffin, Alex Holub, and Pietro Perona.
\newblock Caltech-256 object category dataset.
\newblock 2007.

\bibitem[Cubuk et~al.(2019)Cubuk, Zoph, Mane, Vasudevan, and
  Le]{cubuk2019autoaugment}
Ekin~D Cubuk, Barret Zoph, Dandelion Mane, Vijay Vasudevan, and Quoc~V Le.
\newblock Autoaugment: Learning augmentation strategies from data.
\newblock In \emph{Proceedings of the IEEE/CVF Conference on Computer Vision
  and Pattern Recognition}, pages 113--123, 2019.

\bibitem[xufanxiong(2018)]{xufanxionggithub}
xufanxiong.
\newblock Classification-of-caltech-256, 2018.
\newblock URL
  \url{https://github.com/xufanxiong/Classification-of-CALTECH-256}.

\end{thebibliography}

\end{document}